\documentclass[journal]{IEEEtran}
\ifCLASSINFOpdf
\else
\fi
\usepackage{balance}  
\usepackage{multirow}
\usepackage{graphics}
\usepackage{floatrow}
\usepackage{times}
\usepackage{amsmath}
\usepackage{wrapfig}
\usepackage{multirow}
\usepackage{amsthm}
\usepackage{amsfonts}
\usepackage{leftidx}
\usepackage{indentfirst}
\usepackage{lipsum}
\usepackage{epsfig}
\usepackage{prettyref}
\usepackage{colortbl}
\usepackage[e]{esvect} 
\usepackage{amssymb} 
\usepackage{graphicx}
\usepackage{caption}
\usepackage{subcaption}
\usepackage{algorithm}
\usepackage{algpseudocode}

\theoremstyle{plain}
\theoremstyle{definition}
\usepackage{enumerate}
\newenvironment{packed_enum_i}
{\begin{enumerate}[(i)]
		\setlength{\itemsep}{1pt}
		\setlength{\parskip}{0pt}
		\setlength{\parsep}{0pt}
	}{\end{enumerate}}
\usepackage{arydshln} 
\usepackage{hyperref}
\usepackage[noadjust]{cite}

\usepackage{cases}

\makeatletter
\newcommand*{\algrule}[1][\algorithmicindent]{\makebox[#1][l]{\hspace*{.5em}\vrule height .75\baselineskip depth .25\baselineskip}}%

\newcount\ALG@printindent@tempcnta
\def\ALG@printindent{%
    \ifnum \theALG@nested>0
        \ifx\ALG@text\ALG@x@notext
            \addvspace{-3pt}
        \else
            \unskip
            \ALG@printindent@tempcnta=1
            \loop
                \algrule[\csname ALG@ind@\the\ALG@printindent@tempcnta\endcsname]%
                \advance \ALG@printindent@tempcnta 1
            \ifnum \ALG@printindent@tempcnta<\numexpr\theALG@nested+1\relax
            \repeat
        \fi
    \fi
    }%
\usepackage{etoolbox}
\patchcmd{\ALG@doentity}{\noindent\hskip\ALG@tlm}{\ALG@printindent}{}{\errmessage{failed to patch}}
\makeatother

\usepackage{booktabs}
\usepackage{tikz}
\usepackage{mathrsfs}  

\newcommand{\realfield}{\hbox{I \kern -.4em R}}

\newcommand*{\diameter}{\bigcirc\kern-0.95em\diagup}

\usepackage{leftidx}
\begin{document}
\title{Modal-based Kinematics and Contact Detection of Soft Robots
}
%
%
%

\author{Yue~Chen, Long~Wang, Kevin Galloway, Isuru Godage,
	Nabil~Simaan, and Eric~Barth
\thanks{Y. Chen is with the Department of Mechanical Engineering, University of Arkansas, Fayetteville, AR,72701 USA email: yc039@uark.edu 
	
	L. Wang is with the Department of Mechanical Engineering, Columbia University, NY, 10025 USA email: lw2424@columbia.edu 
	
	I. Godage is with the School of Computing, College of Computing and Digital Media, Chicago, 60604 USA email: igodage@depaul.edu
	
	K. Galloway, N. Simaan, and E. Barth are with the Department of Mechanical Engineering, Vanderbilt University, Nashville,TN, 37235 USA e-mail: {kevin.c.galloway, nabil.simaan, eric.j.barth}@vanderbilt.edu}
}
\maketitle
%
%


\begin{abstract}
Soft robots offer an alternative approach to manipulate inside the constrained space while maintaining the safe interaction with the external environment. Due to its adaptable compliance characteristic, external contact force can easily deform the robot shapes and lead to undesired robot kinematic and dynamic properties. Accurate contact detection and contact position estimation are of critical importance for soft robot modeling, control, trajectory planning, and eventually affect the success of task completion. In this paper, we focus on the study of 1-DoF soft pneumatic bellow bending actuator, which is one of the fundamental components to construct complex, multi-DoF soft robots. This 1-DoF soft robot is modeled through the integral representation of the spacial curve. The direct and instantaneous kinematics are calculated explicitly through a modal method. The fixed centrode deviation (FCD) method is used to to detect the external contact and estimate contact location. Simulation results indicate that the contact location can be accurately estimated by solving a nonlinear least square optimization problem. 


%
%
\end{abstract}
\section{Introduction}
\par Bio-inspired soft robots are robotic systems made of materials that have the similar moduli order of natural organisms ($10^4 - 10^9 \text{Pa}$) \cite{rus2015design}. Different from the conventional rigid-link robotic systems with impedance control \cite{part1985impedance} or stiffness modulation \cite{simaan2003geometric}, compliance is an intrinsic property of soft robots and is achieved through the material mechanical property (for example, the softness of silicone) or morphological control \cite{laschi2014soft}. Soft robots enable safe interaction with external objects and exhibit capability to manipulate inside confined environments or achieve complicated tasks. These robots have been used to grasp objects \cite{ilievski2011soft}, to adapt to environments \cite{morin2012camouflage}, to estimate external shapes \cite{GallowaySoftRobot}, to locomote on rough terrain \cite{tolley2014resilient}, to regain motor skills in rehabilitation applications \cite{polygerinos2015soft}, to support heart function \cite{roche2017soft} and many more \cite{majidi2014soft, kim2013soft, trivedi2008soft}.
\par Soft robots are usually actuated in two forms, namely the variable length tendons and pneumatic/fluidic actuation \cite{rus2015design}. Pneumatically actuated soft robots offer certain advantages due to the benefit of high power-to-weight ratio and easy implementation. Pneumatic muscle actuators (PMA) or McKibben actuator, a simple one degree of freedom (DoF) pneumatic soft actuator fabricated by a rubber tube and fiber sleeves, was proposed for artificial limb research since 1950's \cite{schultecharacteristics} and commercialized in the 1980s by Bridgestone Company. The history of  pneumatically actuated soft robotic system dates back to 1992 when Suzumori et al. developed a 3-DoF (pitch, yaw, and extension) microactuator \cite{suzumori1992applying}. Thanks to the widespread availability of 3D printing technique and affordable fabrication materials, the design and fabrication of pneumatically actuated soft robotics has been continuously growing in the past 10 years. For instance, a pneumatically-actuated soft bending actuator can be fabricated by (1) fabricating the outer molds of the soft actuator, (2) pouring the silicone to the molds, (3) leaving the silicone to cure, and (4) demolding the actuator \cite{softroboticstoolkit}.
\par Forward and inverse kinematic modeling of pneumatic soft robot are required to understand the robot inherent characteristic and enable accurate closed-loop control. In practice, each segment of a soft robot can be regarded as a continuum section when it is powered. Several approaches have been reported to study soft or continuum robot modeling. Piecewise constant-curvature assumption has been widely used in different soft/continuum robot forward kinematics modelings, such as the D-H parameter approach \cite{hannan2003kinematics}, Frenet-Serret frames approach \cite{webster2010design}, exponential coordinates approach \cite{sears2006steerable} (see the detailed continuum robot kinematic modeling in \cite{webster2006nonholonomic}). This assumption provides a simple method to obtain the mapping from robot configuration space to task space. However, the piecewise constant curve assumption is not valid for all the conditions (see Figure 4 of \cite{wakimoto2009miniature} or Figure~\ref{fig:softrobottrialone} in this paper), especially when the shape of a soft or continuum robot is significantly changed due to the large internal joint space pressure input, requiring more accurate modeling and control. A more general approach is a modal representation proposed by Chirikjian et al. in 1994 \cite{chirikjian1994modal}, which described the kinematics of a hyper-redundant snake robot with both constant and variable curvature. Wang and Simaan \cite{wang2019geometric,wang2014investigation} adapted this modal representation for multi-backbone continuum robots and presented kinematic error propagation and geometric calibration to address equilibrium shape deviation. Similar to the rigid serial robot, the inverse kinematics of soft continuum robot is more challenging compared to its forward kinematics. Neppalli et al. developed a  closed-form geometric approach by modeling each segment of a continuum robot as a spherical joint and rigid link and using the conventional analytical method to solve the inverse kinematics \cite{neppalli2009closed}. The Jacobian-based approach is able to search the arc parameters to achieve the desired configuration \cite{sears2006steerable, webster2009closed}. Garbin et al. recently proposed a design that mechanically couples a pneumatic parallel bellows actuator (PBA) with a multi-backbone continuum user interface (CUI), and presented a unified kinematics for both the PBA (soft robot) and the CUI (continuum) \cite{garbin2018dual}. 
\par With the aforementioned kinematic modeling approaches, the soft robots could be accurately controlled in the free space environment. However, due to the compliance of the soft robot, an external wrench disturbance (e.g. force or torque) in a realistic environment could potentially lead to significant robot configuration variation, resulting in inaccurate control. To estimate the external wrench applied to the soft robot body, Ozel et al. proposed a soft robot curvature sensing technique by integrating a resistive flex sensor or a magnetic curvature sensor \cite{ozel2016composite}. Kramer et al. proposed a polydimethylsiloxane (PDMS) based curvature sensing technique in 2011 \cite{kramer2011soft}. However, the soft robot kinematic model variation caused by the external contact remains unsolved.

\par Inspired by the work done by Bajo \cite{bajo2010finding,bajo2012kinematics}, this paper aims (i) to develop a new kinematic modeling approach for soft robot with generalized configurations and (ii) to enable accurate contact detection during motion with feasible sensing feedback. We will systematically study the kinematic modeling of soft robot with contact, contact detection and contact localization. The contact force magnitude estimation is out of the scope of current work. We use an 1-DoF pneumatic bending actuator \cite{wang2016interaction} to experimentally validate the modeling method. 
This paper is a step toward our overarching goal of robust soft robot control method for actual applications when the soft robot is in contact with the external objects. 
The rest of the paper is arranged as follows. Section \ref{sec: PS} details the research problems and the basic modeling assumptions. Section \ref{sc: method} describes the theoretical kinematic modeling and contact detection algorithm. Section \ref{sc:result} presents the simulation results followed by Conclusions in section \ref{sc:conclusions}.
\section{Problem Statement}
\label{sec: PS}
\begin{figure}[h]
	\centering
	\includegraphics[width=0.8\linewidth]{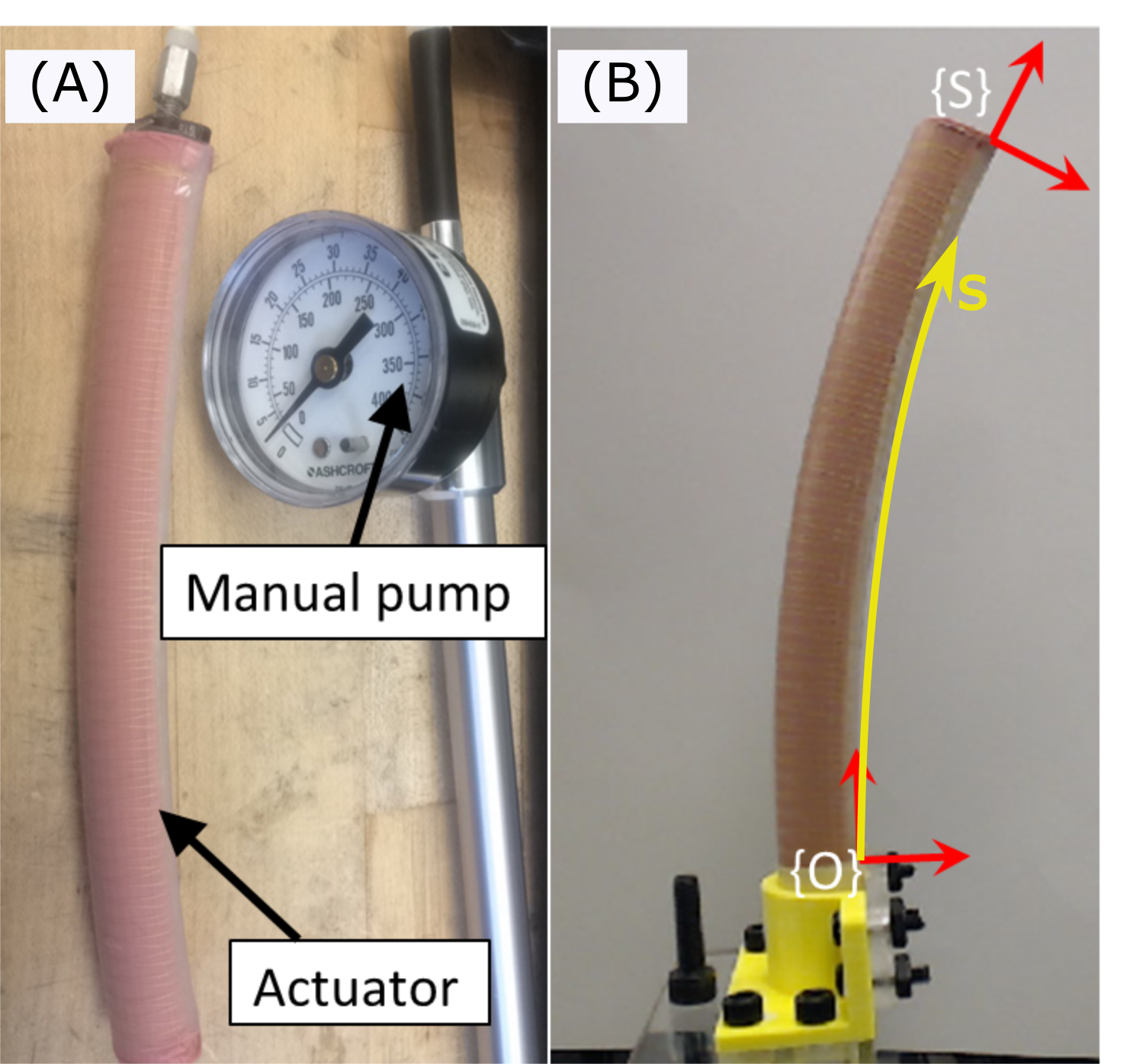}
	\caption{(A) Experimental setup for 1-DoF pneumatic bellow soft actuator; (B) Side view of the pneumatic bellow actuator with 10Psi input pressure provided. Both inertial coordinate frame and end effector coordinate frame are located at the inextensible radial base of the actuator. Arc length $s$ is measured from the bottom to bellow tip with the maximum value of $L$.}
	\label{fig:softrobotfigure}
\end{figure}

Figure \ref{fig:softrobotfigure} shows the soft robot prototype, which consists of a pneumatic bellow soft actuator and a manual pneumatic pump. The actuator fabrication process with fiber reinforcement technique was detailed in \cite{polygerinos2015modeling}. The cross-section of pneumatic bellow actuator is semicircle shape, which has shown to have the smallest bending resistance compared to the rectangular shape and circle shape. A strain-limiting layer was added to its radial base to achieve the uni-directional bending motion and to prevent the axial extending and contracting motion. The circumferential reinforcement wires prevent the pneumatic bellow from significant radial expansion at high joint space pressure input. The maximum bending angle is occurred when the input pressure is 25 psi. 
In this paper, we aim to address the following three problems: 
\begin{packed_enum_i}
	\item[] \underline{\textit{Problem 1}}: Derive the generalized kinematic model of the 1-DoF pneumatic soft bellow actuator. Find the direct kinematics and instantaneous kinematics that maps from robot actuation space to task space.
	\item[] \underline{\textit{Problem 2}}: Given the soft bellow end effector position and velocity, determine whether the external contact is applied to the bellow or not.
	\item[] \underline{\textit{Problem 3}}: Given the soft bellow end effector position and velocity, find the contact location along the backbone of the bellow if the external contact is applied to the robot. 	
\end{packed_enum_i}
\par When discussing these problems, we use the following modeling assumptions:
\begin{packed_enum_i}
	\item The pneumatic bellow bends in the plane. 
	\item The curve of pneumatic bellow and its first-order derivative are continuous.
	\item The pneumatic bellow end effector position and orientation can be obtained from an extrinsic sensor (e.g. electromagnetic tracker).
	\item The portion of a soft bellow from base to contact position maintains its shape after contact and the unconstrained portion continues to bend as if it was a shorter bellow. 
\end{packed_enum_i}
\section{Materials and Methods}
\label{sc: method}
\subsection{Forward Kinematics: Actuator Space to Task Space}
\label{sc: FK}
\begin{figure}
	\centering
	\includegraphics[width=0.99\linewidth]{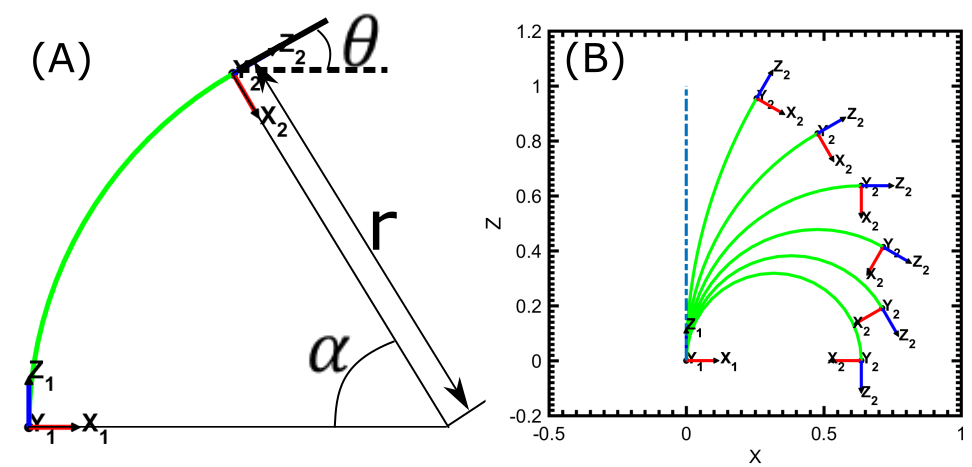}
	\caption{(A)The pneumatic bellow soft actuator bends in the $x-z$ plane with the constant radius of $r$ and bending angle $\alpha$. (B) Expression singularity occurs when the pneumatic is straight}
	\label{fig:constant-curvature-explain}
\end{figure}

The kinematics of continuum robots can be described using two concatenated mappings according to \cite{webster2010design,wang2019geometric}. The first one is a mapping from actuator space (input air pressure $q$ in this paper) to configuration space (tangential angle ${\theta}_s$ along the bending curve at the arc length of $s$ and the given input pressure $q$). The second one is from the configuration space (tangential angle ${\theta}_s$)  to task space (soft actuator tip position and orientation). With the piecewise constant-curvature assumption \cite{jones2006kinematics}, the robot end effector position in the inertial frame can be written as $\mathbf{p}=[r(1-\cos\alpha),\,  0,\, r\sin\alpha]^T$. The end effector orientation can be achieved by rotating about the $y-$axis for $\alpha$ degree (see Figure \ref{fig:constant-curvature-explain}). Therefore, the homogeneous transformation from robot based to end effector can be written in equation \ref{eq:TR} 
\begin{align}
	\mathbf{T} & =\left[\begin{array}{cc}
		R_y(\alpha) & \mathbf{p} \\
		0 & 1 
	\end{array}\right] =\left[\begin{array}{cccc}
		\cos{\kappa}s & 0 & \sin{\kappa}s & \frac{1-\cos{\kappa}s}{\kappa}\\
		0 & 1 & 0 & 0\\
		-\sin{\kappa}s & 0 & \cos{\kappa}s & \frac{\sin{\kappa}s}{\kappa}\\
		0 & 0 & 0 & 1
	\end{array}\right]
	\label{eq:TR}
\end{align}
where $s$ is the arc length and $\kappa = 1/r$ is the curvature. However, this forward kinematics expression is problematic when the curvature is zero. As can be seen from figure \ref {fig:constant-curvature-explain}, zero curvature indicates that the continuum robot is a straight line, while the last column of homogeneous transformation matrix is undefined. 

\par To address the above expression singularity and model the soft robot when the piecewise constant-curvature assumption is insufficiently accurate, we use the modal approach to describe the pneumatic bending characteristics from the actuator space to configuration space. Once the soft pneumatic bellow is powered and at the static position, the geometrically exact configuration only varies with respect to the input pressure and is determined by the minimal energy solution. Experimental calibration is performed with the hardware platform in figure \ref{fig:softrobotfigure} to obtain a family of shapes under the specific pneumatic pressure input. The mapping from actuator space to configuration space can be approximated in the following equation \cite{zhang2013design}: 
\begin{equation}
\boldsymbol{{\theta}}(s,q)={\boldsymbol{\psi}{(s)}^T}A{\boldsymbol{\eta}(q)}
\label{eq:modalform}
\end{equation}
where $\boldsymbol{{\psi}}(s)$ and ${\boldsymbol{\eta}(q)}$ are described in the following modal representations
\begin{equation}
{\boldsymbol{\psi}(s)}=[1, s, s^2, \ldots, s^{v-1}]^T
\label{eq:psiRepresentation}
\end{equation}
\begin{equation}
{\boldsymbol{\eta}(q)}=[1, q, q^2,\ldots, q^{w-1}]^T
\label{eq:etaRepresentation}
\end{equation}
The bending actuator backbone can be decomposed into $g$ discrete points along the strain limiting layer. Each point has a specific corresponding tangential angle at $z$ different input pressure. Then equation \ref{eq:modalform} can be expressed in the following matrix form:
\begin{equation}
{\Phi}={\Omega_{g\times{v}}}{A_{v\times{w}}}{\Gamma_{w\times{z}}}
\label{eq:modalmatrixform}
\end{equation}
where ${\Phi_{i,j}}={\theta}(s_i,q_j)$, $i$ is the number of discrete points along the bending actuator backbone and $j$ is the number of input pressure.  $\Omega$ and $\Gamma$ are written as follows
\begin{align}
\Omega & =\left[\begin{array}{cccc}
1 & s=0 & \ldots & s=0^{v-1}\\
1 & s=s_1 & \ldots & s={s_1}^{v-1}\\
\vdots  & \vdots  & \ddots  & \vdots \\
1 & s=s_{max} & \ldots & s={s_{max}}^{v-1} 
\end{array}\right]_{g\times{v}}
\label{eq:Omega}
\end{align}

\begin{align}
\Gamma & =\left[\begin{array}{cccc}
1  & 1 & \ldots & 1\\
q=q_1&  q=q_2 & \ldots  & q=q_{max} \\
\vdots& \vdots  & \ddots  &  \vdots\\
q={q_1}^{w-1} & q={q_2}^{w-1} & \ldots & q={q_{max}}^{w-1} 
\end{array}\right]_{w\times{z}}
\label{eq:Gamma}
\end{align}


Equation \ref{eq:modalmatrixform} can be reshaped into the following form:
 \begin{equation}
 [{\Phi}^T\otimes \Omega] \mathbf{A}=\mathbf{\Phi}
 \label{eq:kronecker}
 \end{equation}
where $\otimes$ is the Kronecker's product, $\mathbf{A}$ and $\mathbf{\Phi}$ are the vectorized form of $A$ and $\phi$ respectively, which can be written as:
 \begin{equation}
\mathbf{A} = [a_{11} \ldots a_{v1}, a_{12} \ldots a_{v2},\ldots , a_{1w} \ldots a_{vw}]^T
\label{eq:vectorA}
\end{equation}

 \begin{equation}
\mathbf{\Phi} = [\theta_{11} \ldots \theta_{g1}, \theta_{12} \ldots \theta_{g2},\ldots , \theta_{1z} \ldots \theta_{gz}]^T
\label{eq:vectorphi}
\end{equation}

Solving matrix $\mathbf{A}$ could get the mapping from the soft actuator joint space to the configuration space. 
The mapping from  configuration space to task space can be explained by using the definition of curvature (see figure \ref{fig:constant-curvature-explain}), as described in the following equation: 
\begin{equation}
\frac{d\mathbf{x}_s}{ds}=[\cos{\theta_s}\ \ 0 \ \ \sin{\theta_s} ]^T
\label{eq:theta2EE}
\end{equation}
where $\mathbf{x}_s$ is the point position expressed in the inertial frame at the arc length of $s$ and $\theta_s$ is the corresponding tangential angle $s$ (figure \ref{fig:constant-curvature-explain}). 
The forward kinematics that maps from actuator space and task space is obtained by substituting equation \ref{eq:modalform} into equation \ref{eq:theta2EE} and integrate the tangent vector along the arc length.

\subsection{Instantaneous Kinematics }
\label{sub: IK}
This section describes the instantaneous kinematics modeling that maps the actuator speed $\dot{q}$ to the end effector twist  $\mathbf{\dot{x}}=[\boldsymbol{v}^T, \ \ {\boldsymbol{\omega}}^T]^T$. The end effector twist consists of the linear velocity $\boldsymbol{v}$ and angular velocity $\boldsymbol{\omega}$. However, since the actuator bends in the $x-z$ plane, the linear velocity in $y$ direction and angular velocity about $x$ and $z$ axis are zero. We seek to find the Jacobian $J_{xp}$ that satisfies the following equation:
\begin{equation}
\mathbf{\dot{x}} = J_{xq} \dot{q}
\label{eq:x2q_init}
\end{equation}

The Jacobian the relates the joint space motion rates $\dot{q}$ with the end effector twist can be derived by using the differential chain rule: 
\begin{align}
	J_{\mathbf{x}{q}} & = \frac{\partial{\mathbf{x}}}{\partial{q}} \nonumber \\
	&= \frac{\partial{}}{\partial{q}} {\int_{0}^{L}[\cos{\theta_s}\ \ 0\ \ \sin{\theta_s} ]ds } \nonumber \\
	& =  \int_{0}^{L}[\frac{\partial{}}{\partial{q}} \cos{\theta_s}\ \ 0\ \ \frac{\partial{}}{\partial{q}} \sin{\theta_s} ]ds   
	\label{eq:J_x}
\end{align}
where $\frac{\partial{}}{\partial{q}} \cos(\theta_s)$ and $\frac{\partial{}}{\partial{q}} \sin(\theta_s)$ can be written as follows: 

\begin{align}
	\frac{\partial{}}{\partial{q}} \cos(\theta_s) & = -\sin(\theta_s)\frac{\partial{({{\psi(s)}^T}A{\eta{(q)}})}}{\partial{q}}\nonumber \\
	&= -\sin(\theta_s){{{\psi(s)}^T}A \frac{\partial{({\eta{(q)}})}}{\partial{q}}} 
	\label{eq:dcos}
\end{align}


\begin{align}
	\frac{\partial{}}{\partial{q}} \sin(\theta_s) & = \cos(\theta_s)\frac{\partial{({{\psi(s)}^T}A{\eta{(q)}})}}{\partial{q}}\nonumber \\
	&= \cos(\theta_s){{{\psi(s)}^T}A \frac{\partial{({\eta{(q)}})}}{\partial{q}}}  
	\label{eq:dsin}
\end{align}


Combining equation \ref{eq:J_x}, \ref{eq:dcos} and \ref{eq:dsin}, we can get the Jacobian that relates the actuator space speed to the robot end effector twist (linear and angular velocities). Given the specific arc length $L$, the end effector twist is the function of input pressure $q$ and pressure change rate $\dot{q}$. The Jacobian was also validated numerically by using the resolved rates method \cite{whitney1969resolved}, as can be seen in the following equation: 

\begin{equation}
q_i = q_{i-1}+ J^{-1} {\alpha(\mathbf{x}_{des} - \mathbf{x}_{i})}
\label{eq: resolved}
\end{equation}
where $\alpha$ is the positive scaling factor.

\subsection{Contact Detection}
This section describes the contact detection algorithms as initially discussed in \cite{bajo2010finding}. The joint force deviation (JFD) method calculates the external wrench value based on the deviation of generalized joint space forces with respect to the nominal model. It can be described in the following equation: 
\begin{equation}
{J_{q\theta}}^T \boldsymbol{\tau} =\nabla(E)-{J_{x\theta}}^T \boldsymbol{W_e}
\end{equation}
where $\nabla(E)$ is the gradient of elastic energy with respect to the configuration perturbation, $\boldsymbol{\tau}$ and $\boldsymbol{W_e}$ are the joint force and external wrench respectively. When there is no external wrench applied on the soft robot, then ${J_{q\theta}}^T \boldsymbol{\tau} =\nabla(E)$. Once the wrench applied on the robot ($\boldsymbol{W_e} \ne 0 $), which leads to the variation of the joint force if the pre-determined detection threshold is $\xi=0$. Based on the deviation value, we can obtain the external wrench according to the equation below: 
\begin{equation}
\boldsymbol{W_e} = -({{J_{x\theta}}^T})^{-1}{J_{q\theta}}^T\Delta{\boldsymbol{\tau}}
\end{equation}
where $\Delta{\boldsymbol{\tau}}$ is the deviation of joint force. In the real applications, the contact detection threshold $\xi$ is typically not equal to 0, which is caused by the friction force in the pull-wire continuum robots and the compressibility characteristic of pneumatic supply
in the current robot. To implement this algorithm, one needs to have a accurate joint space pressure sensor to get the generalized joint force feedback. However, the major limitation of this algorithm is that it does not provide the contact location \cite{bajo2010finding}.  

The fixed centrode deviation (FCD) method calculates the deviation of theoretical and actual loci of the fixed centrode of end effector to detect the contact position. The loci of theoretical fixed centrode (or instantaneous screw axis, ISA) can be expressed in the following equation \cite{angeles2013fundamentals}:
\begin{equation}
\boldsymbol{c}_m=\frac{\Omega_m(\dot{\mathbf{P}}_m - \Omega_m \mathbf{P}_m)}{{\boldsymbol{\omega}_m}^T\boldsymbol{\omega}_m}
\label{eq:FCmodel}
\end{equation} 
where $\dot{\mathbf{P}}_m$ and $\boldsymbol{\omega}_m$ are the end effector linear velocity and angular velocity respectively, which is calculated from the instantaneous kinematics model described in section \ref{sub: IK}. $\Omega_m$ is the skew-symmetric angular velocity matrix defined as $\Omega_m=[\boldsymbol{\omega}_m]^{\wedge}$ \cite{murray1994mathematical}. Based on the modeling assumption (3) in section \ref{sec: PS}, the actual end effector position and orientation can be captured through a tracking system (e.g. EM tracker or motion camera). The loci of actual fixed centrode is expressed as following:
\begin{equation}
\boldsymbol{c}_s=\frac{\Omega_s(\dot{\mathbf{P}}_s - \Omega_s \mathbf{P}_s)}{{\boldsymbol{\omega}_s}^T\boldsymbol{\omega}_s}
\label{eq:actualFCD}
\end{equation} 
where the subscript $s$ in equation \ref{eq:actualFCD} indicates the sensor data. Once the deviation of fixed centrode occurs, there exists an external wrench along the pneumatic bellow soft robot. Similarly, the detectability threshold $\xi$ also exists in the FCD method and it is affected by the tracking system resolution.

\subsection{Contact Location Estimation}
\label{subsec: Contactlocationestimation}
This section presents the contact location estimation algorithm by using the kinematics modeling and contact detection method described above. Once the contact occurs at a given location $s=s_c$, the robot kinematic model can be decomposed into two sections (see figure \ref{fig:motionwithcontactwithoutinitial}. Note that, without future explanation, $x-$position and $z-$position are expressed in the unit of pixel, which is 0.2959 mm/pixel). Due to the inherent compliance characteristic of the soft robot, the soft robot base to contact position forms the constrained portion. The contact position to end effector forms the unconstrained portion, which continues to bend with the increase joint space pressure. The unconstrained portion kinematic model is similar to the model described in section \ref{sc: FK} and \ref{sub: IK}, which has a shorter arc length and different base position \cite{bajo2010finding} (which is the contact location). The modified kinematic model can be written as:
 \begin{figure}[ht]
 	\centering
 	\includegraphics[width=0.9\linewidth]{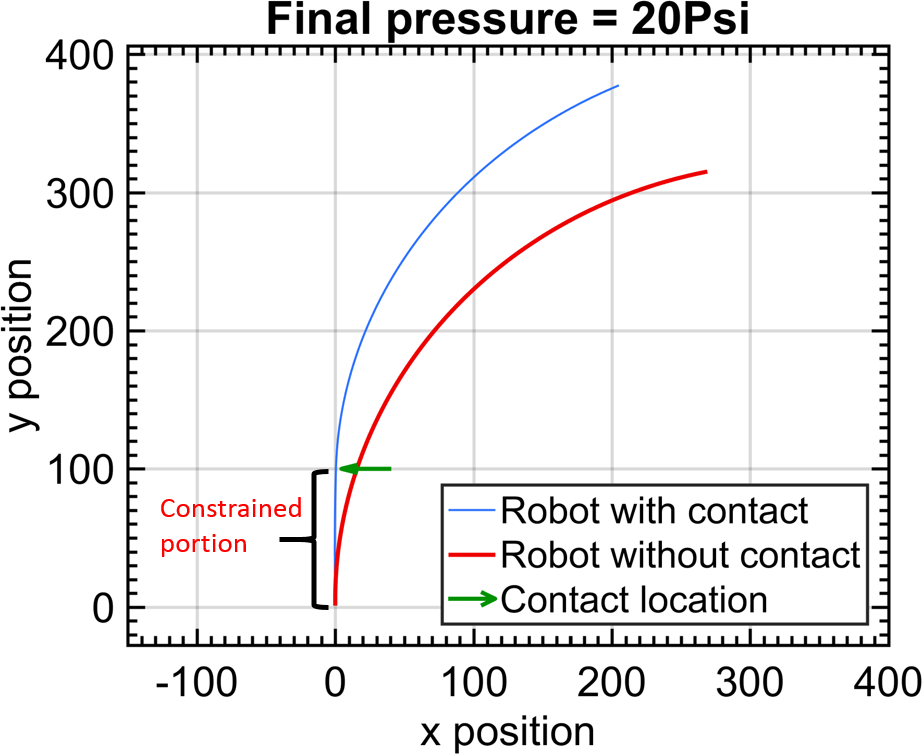}
 	\caption{Contact occurs on the pneumatic bellow actuator at $s=100$ when the input pressure is 5 Psi. The final pressure is 20 Psi.}
 	\label{fig:motionwithcontactwithoutinitial}
 \end{figure}
 \begin{numcases}{{\tilde{{\theta}}(s,q)}=}
 	{\theta_c}(s,q_c) & $0<s\leq s_c$ \label{positive-subnum}
 	\label{eq:constrainedportion}
 	\\
 	{{\psi(s-s_c)}^T}A{\eta(q)} & $s_c<s\leq L$ \label{negative-subnum}
 	\label{eq: unconstrainedportion}
 \end{numcases}
where $q_c$ is the input pressure when contact occurs, $s_c$ is the contact location, and ${\theta_c}(s,q_c), s\in [s_c, L]$ is the tangential angles along the arc when contact occurs. 

Having this relationship, the pneumatic bellow actuator instantaneous kinematics can be obtained readily by changing the actual arc length.
\begin{equation}
\mathbf{\dot{\tilde{x}}} = \tilde{J_{xq}} \dot{q}
\label{eq:x2q}
\end{equation}  
where $\tilde{J_{xq}}$ is the Jacobian of the unconstrained portion by substituting the arc length $s = L-s_c$ into equation \ref{eq:J_x}, \ref{eq:dcos} and \ref{eq:dsin}. Then the loci of the fixed center can be derived by combining equations \ref{eq:theta2EE}, \ref{eq:x2q_init}, \ref{eq:FCmodel}, \ref{eq:constrainedportion},  \ref{eq: unconstrainedportion}, and  \ref{eq:x2q}. 

The contact location is estimated by solving a least square optimization problem that minimizes the Euclidian distance between the actual loci of fixed centrode obtained from sensor data and theoretical loci of the fixed centrode obtained from modeling results. The objective function is given by 
\begin{equation}
\textnormal{argmin}\frac{1}{2}{\boldsymbol{c}_{err}}^TW \boldsymbol{c}_{err}
\end{equation}
where ${\boldsymbol{c}_{err}} = \boldsymbol{c}_s-\boldsymbol{c}_m$ and $W$ is the weighting matrix. The gradient of fixed centrode $c_m$ in terms of contact location $s_c$ needs to be derived to implement this algorithm. According to equation \ref{eq:FCmodel}, the gradient can be written in the following form:
\begin{align}
\frac{\partial \boldsymbol{c}_m}{\partial s_c}=\frac{\partial \boldsymbol{c}_m}{\partial \boldsymbol{\omega}_m}\frac{\partial \boldsymbol{\omega}_m}{\partial s_c}+ \frac{\partial \boldsymbol{c}_m}{\partial \dot{\mathbf{P}}_m}\frac{\partial \dot{\mathbf{P}}_m}{\partial s_c}+\frac{\partial \boldsymbol{c}_m}{\partial {\mathbf{P}}_m}\frac{\partial {\mathbf{P}}_m}{\partial s_c}+\frac{\partial \boldsymbol{c}_m}{\partial \boldsymbol{\omega}_m}\frac{\partial \boldsymbol{\omega}_m}{\partial s_c}
\end{align}

With the gradient calculated above, the least square optimization problem will be solved iteratively according to \cite{lancaster1985theory}.

\section{Results}
\label{sc:result}
This section describes the results of pneumatic bellow calibration performance, forward/instantaneous kinematics, motion simulation after contact, and contact position estimation. 
\begin{figure*}[!t]
	\centering
	\includegraphics[width=0.8\linewidth]{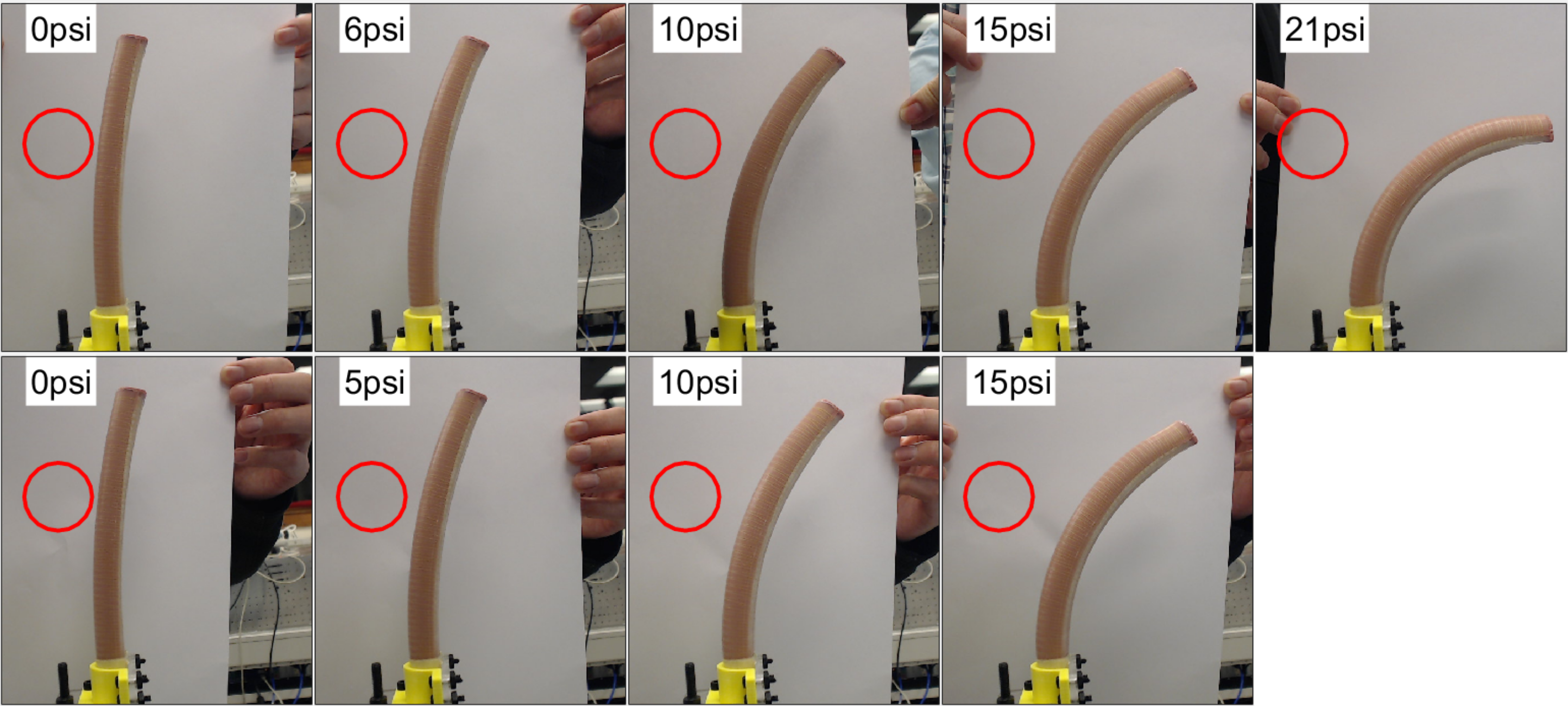}
	\caption{Pneumatic bellow actuator shape with respect to pressure input. The first row indicates that the input pressure increase from 0 to 21 Psi; the second row indicates the input pressure from 21 Psi to 0.}
	\label{fig:softrobottrialone}
\end{figure*}
\subsection{Pneumatic Bellow Calibration}
\label{subsc: calibration}
This section presents the pneumatic bellow calibration results. The experimental setup includes a soft bending bellow, manual pneumatic pump, a high resolution camera (C920 Webcam, Logitech, Swiss) and custom designed graph user interface (GUI). The manual pneumatic pump has the resolution of 1Psi. High resolution input pressure control can be achieved by closing the pressure control loop with a digital pressure sensor. The pneumatic bellow with different air pressure inputs was captured by the camera system and stored for post processing. In this experiment, the input pressure samples 0 Psi, 6 psi, 10 Psi, 15 Psi and 21 Psi respectively  (Figure \ref{fig:softrobottrialone}).
\par
\begin{figure}[!b]
	\centering
	\includegraphics[width=0.9\linewidth]{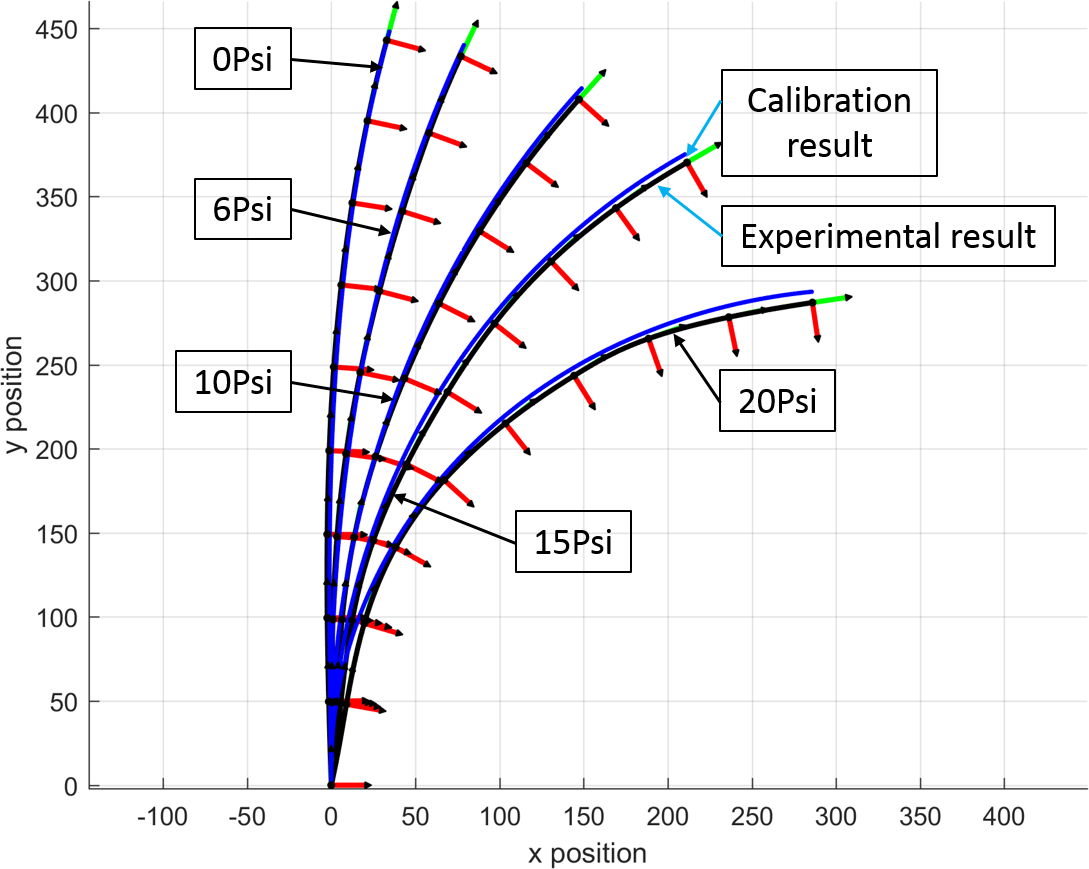}
	\caption{The blue curves show the calibration result. The black curves show the pneumatic bellow actuator with the input pressure of 0 psi, 6 psi, 10 psi, 15 psi and 21 psi respectively. The coordinates frames along the arc indicate the point used for modal fitting.}
	\label{fig:modalanalysisv2}
\end{figure}

 The strain limiting layer on the pneumatic bellow was manually segmented to represent the pneumatic bellow shape. We annotated ten points along the each backbone and find the corresponding positions and tangential angles. By solving equation \ref{eq:kronecker}, we can obtain the model-dependent $\mathbf{A}$ matrix. The segmented curve and calibrated model can be seen in Figure~\ref{fig:modalanalysisv2}. The result shows that the calibrated model closely matches the experimental results. The difference between the calibrated model and the actual shape increases with the increase of input pressure. However, the maximum error occurred at 21 Psi input pressure is less than 2.1 mm. Increasing the number of point $g$ could increase the calibration accuracy at the expense of increasing the computation load. 
%

\subsection{Instantaneous Kinematics Evaluation}
\label{subsc: instantaneouskinematics}
\begin{figure*}[t]
	\centering
	\includegraphics[width=0.99\linewidth]{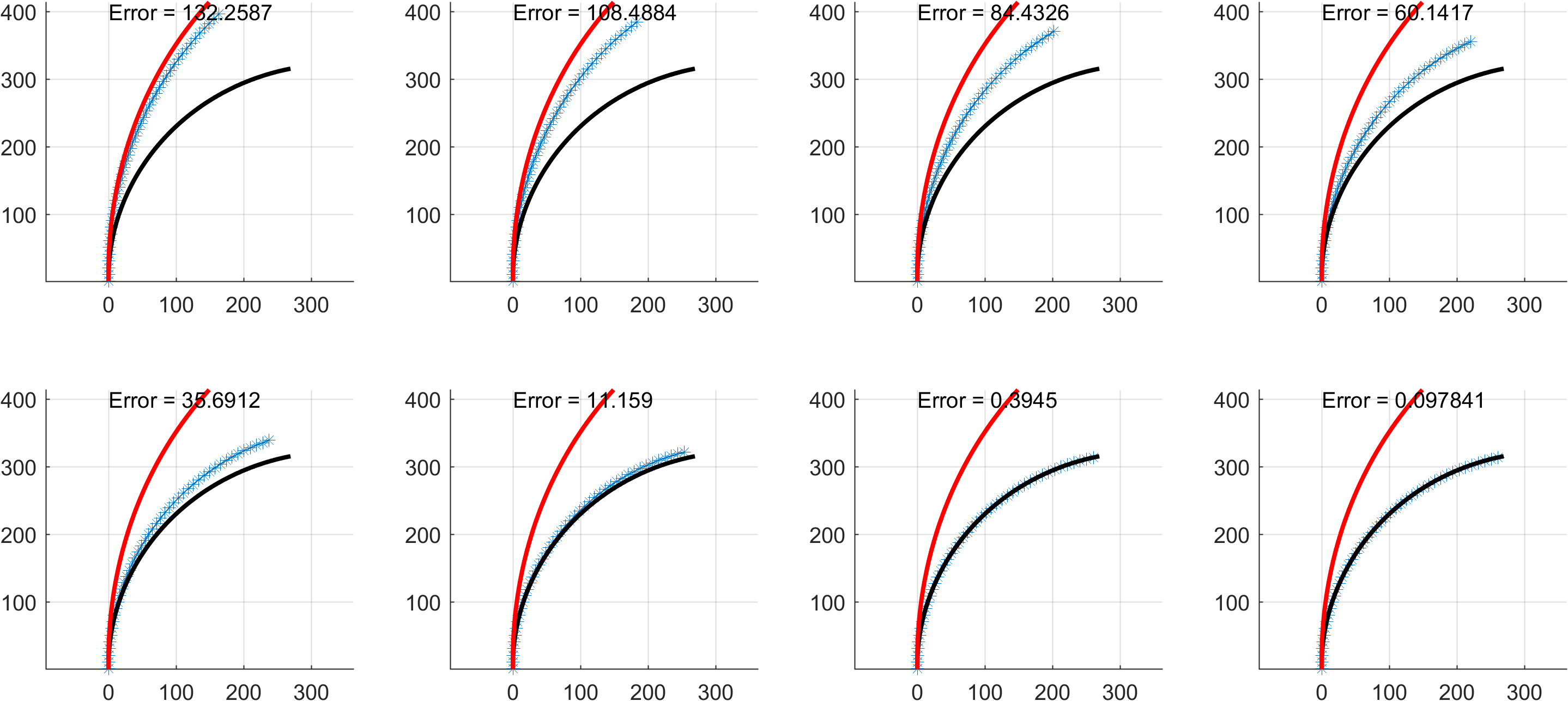}
	\caption{Pneumatic bellow actuator instantaneous kinematics model validation. The red solid curve indicates the actuator  initial status; black solid curve indicates the actuator final status; the curve with asterisk indicates the process when the pneumatic bellow actuator converges to the desired status}
	\label{fig:instantaneouskinematicsverification}
\end{figure*}
This section validates the instantaneous model of the pneumatic bellow actuator as presented in equation (\ref{eq: resolved}). Given the actuator initial position and desired position, the resolved rates algorithm is applied to find the corresponding joint space pressure input. Note that the instantaneous kinematics model is evaluated in the scenario of no contact occurs. Figure \ref{fig:instantaneouskinematicsverification} shows the iterative process when the pneumatic bellow converges to the desired position. The pneumatic bellow actuator end effector space error is 0.0326 mm (0.0978 pixels) and the joint space error is 0.016 Psi, which validates the accuracy of the proposed instantaneous kinematics model.   

\subsection{Contact Detection and Estimation}
\label{subsc: contactdetection}
\begin{figure}[h]
	\centering
	\includegraphics[width=0.99\linewidth]{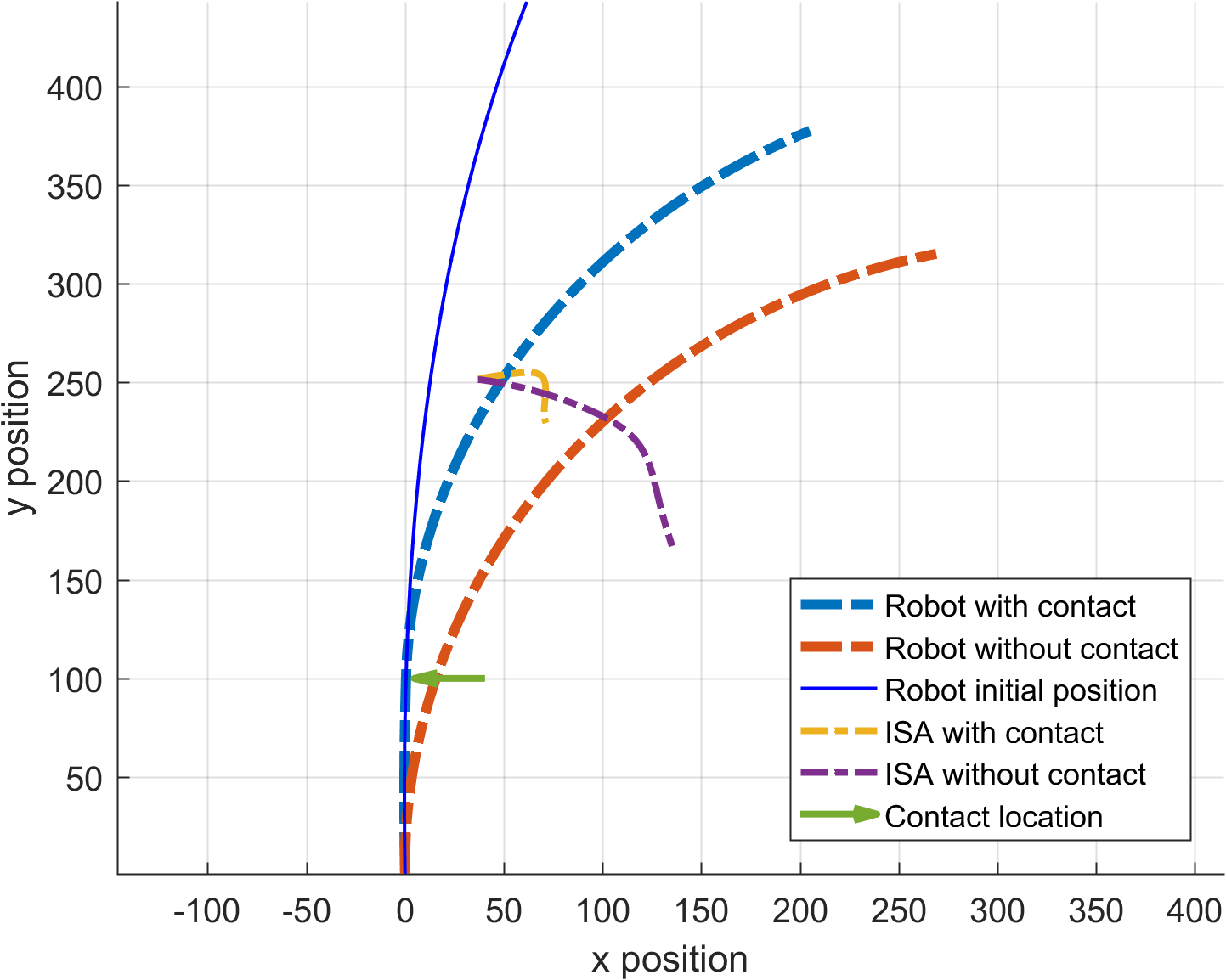}
	\caption{Robot motion with and without contact. The contact occurs at $\text s=100$ when input pressure is 5 Psi (green arrow) with the pressure increase rate of 0.05 Psi/step and robot motion stops when the input pressure is 20 Psi. ISA for both cases from 5 Psi to 20 Psi are shown in this figure.}
	\label{fig:motionwithcontact}
\end{figure}

This section presents the contact detection and contact location estimation. In the experimental study, the contact occurs at $\text s_c = 100$. The input pressure was increased from 5 Psi to 20 Psi with the step size of 0.05 Psi. The loci of fixed centrode throughout the trajectory was calculated simultaneously. We also provided the robot motion without any contact for comparative study. The simulation results are shown in Figure~\ref{fig:motionwithcontact}, which illustrates that the ISA difference between these two scenarios increases steadily due to the growth of input pressure. This is mainly due to the significant end effector position difference as the pressure goes up, see equation (\ref{eq:FCmodel}). Maximum ISA difference is observed when the input pressure is 20 Psi. 
\par
\begin{figure}
	\centering
	\includegraphics[width=0.99\linewidth]{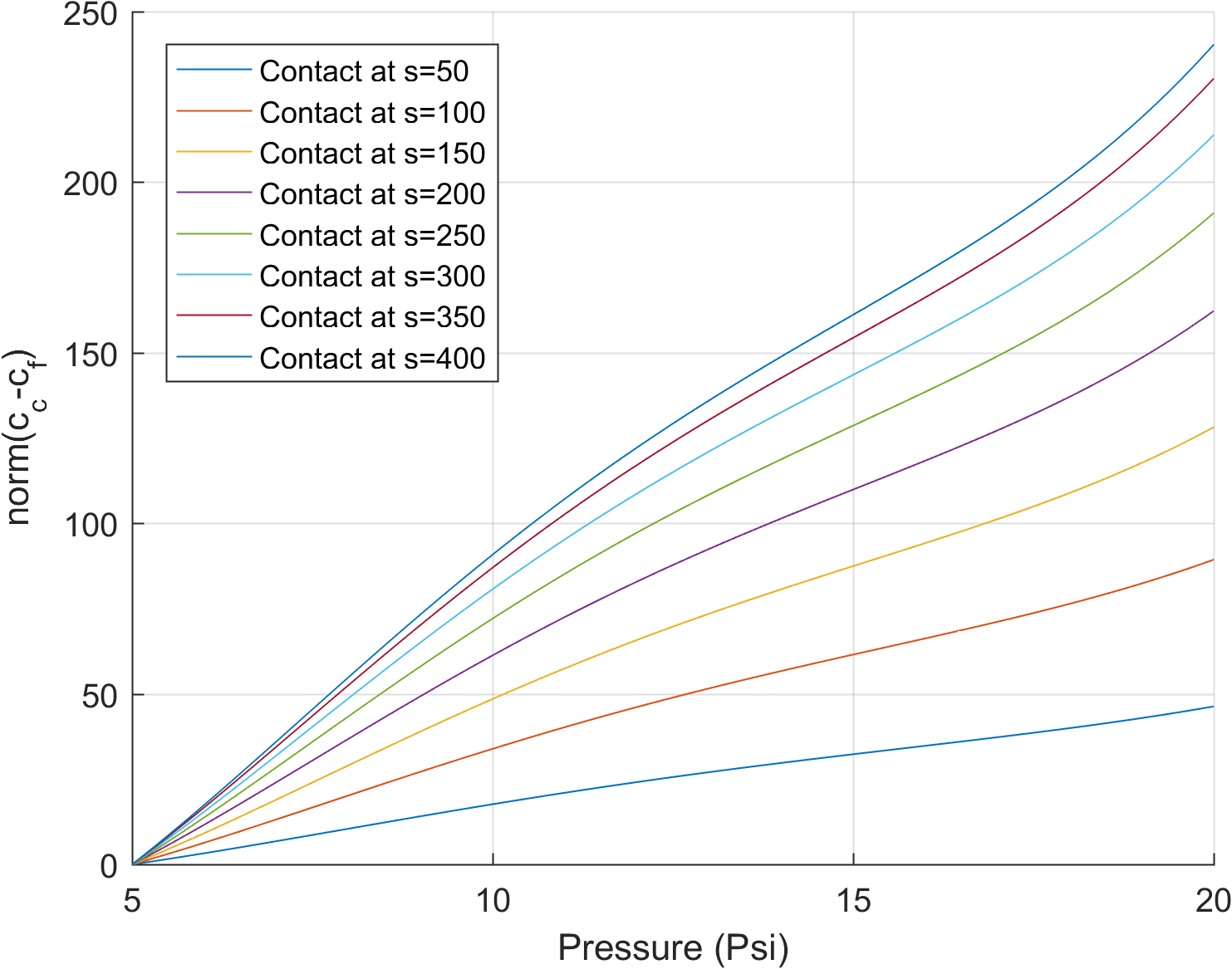}
	\caption{ISA difference affected by the joint space pressure input and external contact location}
	\label{fig:isahistory}
\end{figure}

 We also simulate the relation between ISA difference vs. the contact location. In the simulation, the contact position was chosen at $\text s_c=50$ to 400 from the actuator base with 50 increment per step (Figure~\ref{fig:isahistory}). The performance index is defined as follows:
\begin{equation}
\text {norm}(\boldsymbol{c}_c-\boldsymbol{c}_f)
\end{equation}
where $\boldsymbol{c}_c$ is the fixed centrode when robot is in contact and $\boldsymbol{c}_f$ is the fixed centrode when robot is free of contact. The ISA difference serves the sensitivity performance for the FCD method and it rises as a result of the contact location moves further from the base. Intuitively, contacts can be detected easily when they are closer to end effector due to the large difference in both end effector twist and position when such contacts occur. In contrast, it is difficult to detect contacts when they are close to the base. The ISA difference equals to 0 when $\text s_c=0$ (indicates that the contact point is located at the fixed base of the soft bellow actuator), which also represents that the contact does not affect the robot shape and the ISA. 


\par
\begin{figure}[t]
	\centering
	\includegraphics[width=0.99\linewidth]{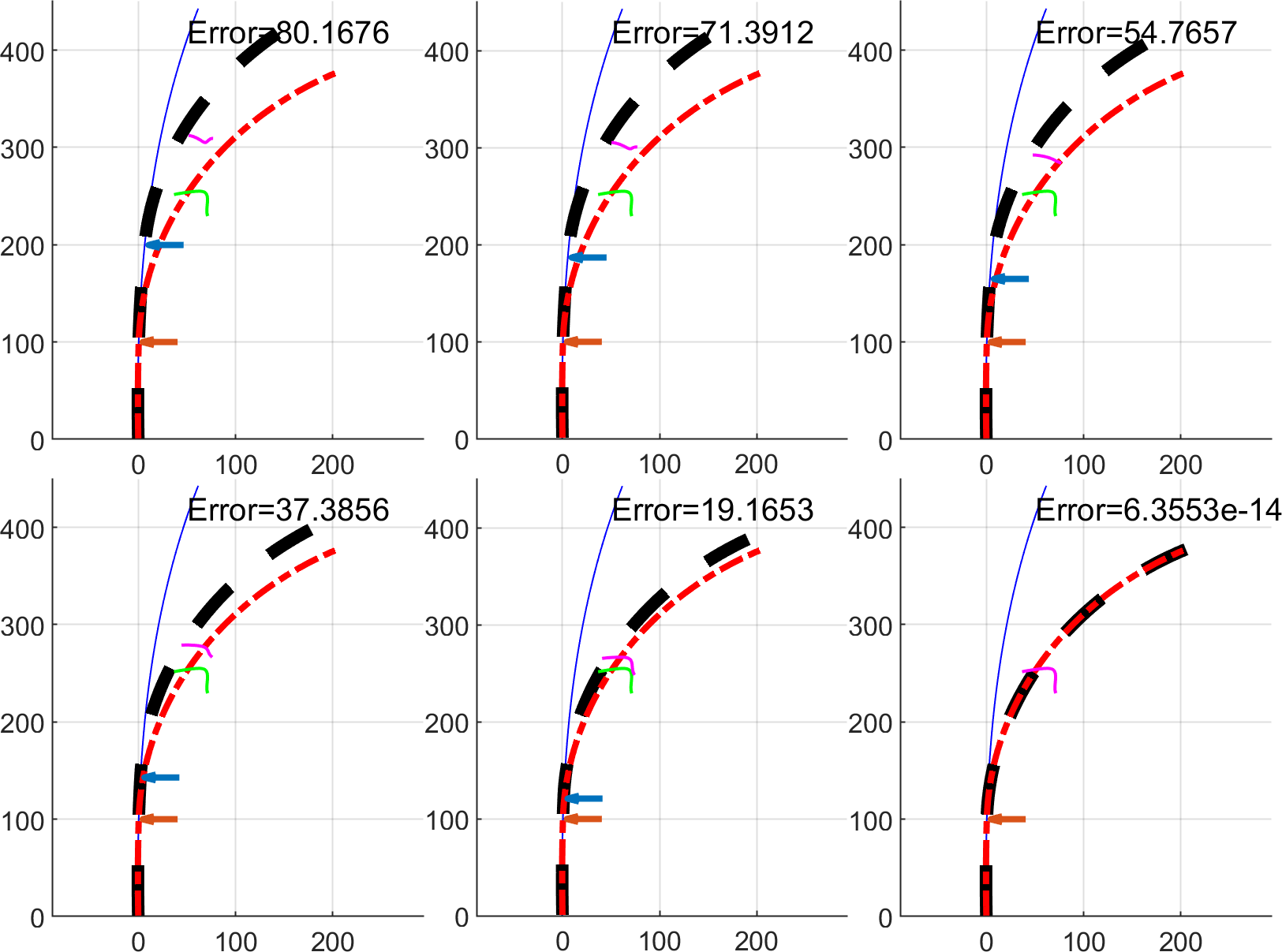}
	\caption{Contact position estimation convergence process with the initial guess of $\text s = 200$. The thin blue curve indicates the starting status; black thick dashed curve indicates the pneumatic actuator with estimated contact location; red thin dashed curve indicates the pneumatic actuator with actual contact location. The ISA for both cases are shown in the magenta and green color. The arrows are the estimated and actual contact position. The algorithm successfully converge to the actual position with the task space error of zero. }
	\label{fig:contactestimationconvergence}
\end{figure}

\begin{figure}[t]
	\centering
	\includegraphics[width=0.99\linewidth]{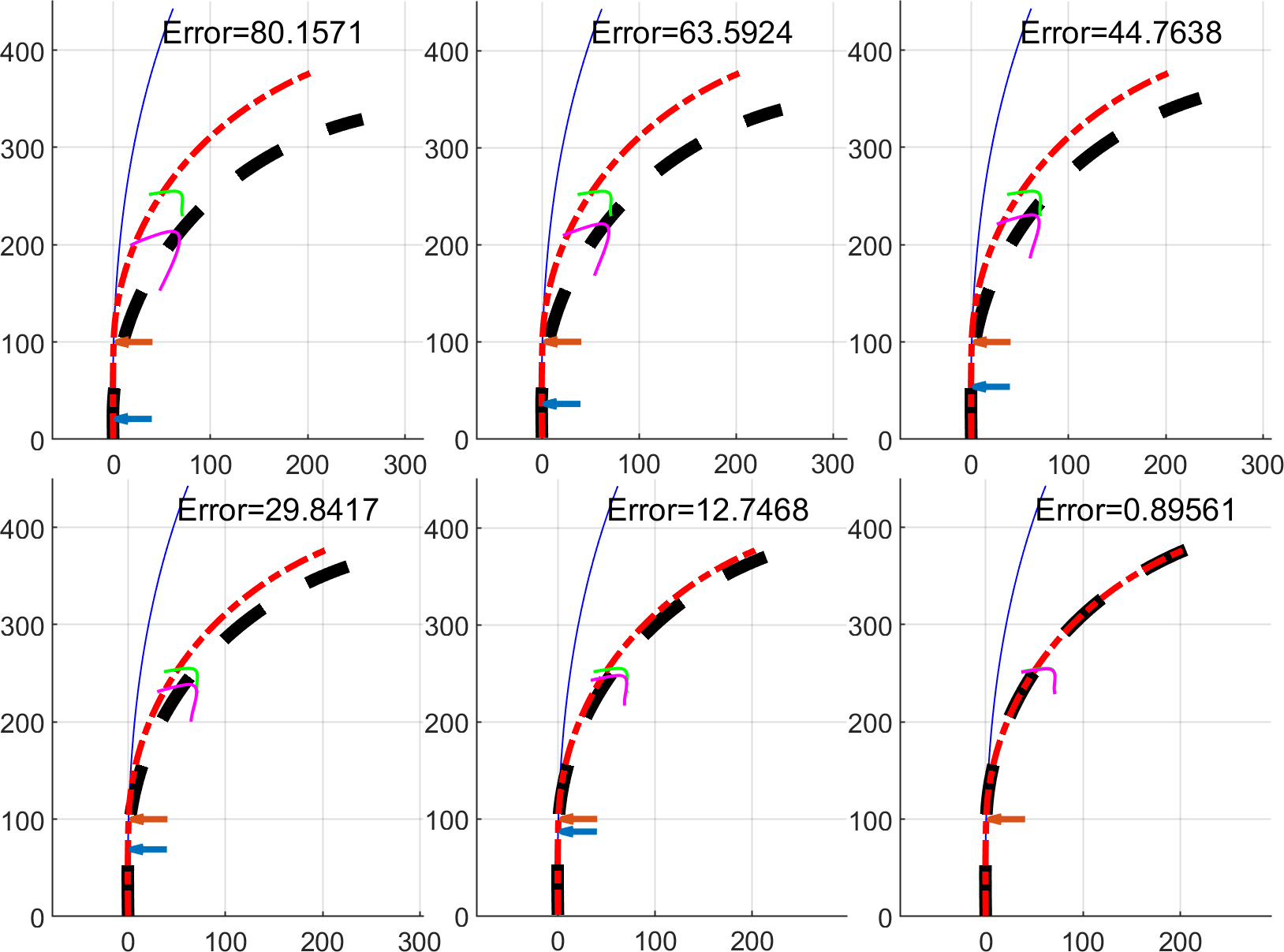}
	\caption{Contact position estimation convergence process with the initial guess of $\text s = 20$. The algorithm successfully converge to the actual position with the task space error of 0.89 pixels.}
	\label{fig:contactestimationconvergencebottom2top}
\end{figure}

 Next we present the contact position estimation result by using the method discussed in section \ref{subsec: Contactlocationestimation}. The ISA with a given contact location ($\text s_c = 100$) is obtained from the simulation results and used as the ground truth. The estimation process starts with an initial guess ($\text s_c = 200$) and then calculates the ISA error as well as the $\frac{\partial \boldsymbol{c}_m}{\partial s_c}$. Using the nonlinear least square optimization method, the solution iteratively converge to the ground truth. Another initial guess ($\text s_c = 20$) was also tested. Both cases studies were able to find the contact position with the acceptable accuracy (figure \ref{fig:contactestimationconvergence} and \ref{fig:contactestimationconvergencebottom2top}).

\section{Conclusions}
\label{sc:conclusions}
This paper presents the a mathematical framework for kinematic modeling, contact detection, and contact location estimation of a 1-DoF soft bellow actuator. A modal representation approach was used to derive the forward and instantaneous kinematics model.  We presented the simulation study of the pneumatic bellow actuator with contact by using a modified kinematic model. 
The fixed centrode deviation was used as the performance index for contact detection. Contact position was estimated by solving a nonlinear least square optimization problem. Simulation results show that the proposed method could accurately estimate the contact location with sub-pixel error. 

The results in this paper show for the first time how the mapping from joint space and task space can be generated through the modal approach and how the contact detection could be estimated through a least square optimization manner. This will inspire future development of complex soft robot modeling and contact control to accomplish the desired tasks. 
\bibliographystyle{IEEEtran}
\bibliography{ms}
\newpage
%
%
%
%
%
%
%
%

\end{document}